\documentclass[runningheads]{llncs}
\usepackage[T1]{fontenc}
\usepackage{graphicx}
\usepackage{wrapfig}
\usepackage{subcaption}
\usepackage{amsmath}
\usepackage{amsfonts}
\newcommand{\printfnsymbol}[1]{%
  \textsuperscript{\@fnsymbol{#1}}%
}

\begin{document}
\title{Spatial-Temporal Expert Learning for Video-based Person Re-identification}
%
%
\author{Xiaofei Hui\inst{1}, Pengfei Wang\inst{1}, Evan Ling\inst{2}, Dezhao Huang\inst{2}, Keng Teck Ma\inst{2}, Minhoe Hur\inst{2}, Jun Liu\inst{1}\textsuperscript{†}}
\authorrunning{X. Hui et al.}
%
\institute{Singapore University of Technology and Design, Singapore \and Hyundai Motor Group Innovation Center in Singapore (HMGICS), Singapore 
}
\def\thefootnote{{\textsuperscript{†}}}\footnotetext{Corresponding Author.}
\maketitle              
\begin{abstract}
Video-based person re-identification (Re-ID) aims to retrieve the same identity in the query video clips from the gallery video clips. To solve this problem, exploiting fine-grained features is of great importance, especially when discriminating identities that are similar in appearance. In this paper, we propose to enhance the ability to explore fine-grained information with a novel input-aware extendable expert module. Instead of updating the network parameters with every sample in the dataset, we aim to train the experts within specific subsets that only contain similar samples and promote their ability to exploit fine-grained information within these similar samples. To achieve this goal, we incorporate two mechanisms in this module: input-aware expert selection mechanism and spatial-temporal selection mechanism.
The first mechanism dynamically activates a set of experts on subsets of similar samples, pushing the experts to exploit subtle differences between these similar samples, while the second one further increases their sensitivity to the fine-grained differences in spatial and temporal aspects and allows the experts to dynamically utilize them for different input samples. In addition, to facilitate the expert module, we design an extendable scheme that allows the module to flexibly add new experts when necessary. As a result, our method achieves outstanding performance on two large-scale datasets.

\end{abstract}
\section{Introduction}
\label{sec:intro}

Given a query image or video, person re-identification (Re-ID) aims to determine whether the same identity has appeared in other cameras or in the same camera at a different time. 
As a core capability for intelligent video surveillance and smart city applications, it has attracted sustained attention in recent years. Compared with image-based Re-ID, video-based Re-ID can exploit richer appearance and temporal cues, making it more robust to occlusion, viewpoint change, and missing body parts~\cite{bai2022salient,DenseIL,BiCnet_TKS,yang2020spatial,wang2021pyramid}. Despite the promising progress of existing methods~\cite{bai2022salient,DenseIL,wang2021pyramid,liu2021watching,aich2021factorization}, video-based Re-ID remains challenging in real-world surveillance scenarios.

\begin{wrapfigure}{r}{0.5\linewidth}
\vspace{-5mm}
  \centering
  \includegraphics[width=\linewidth]{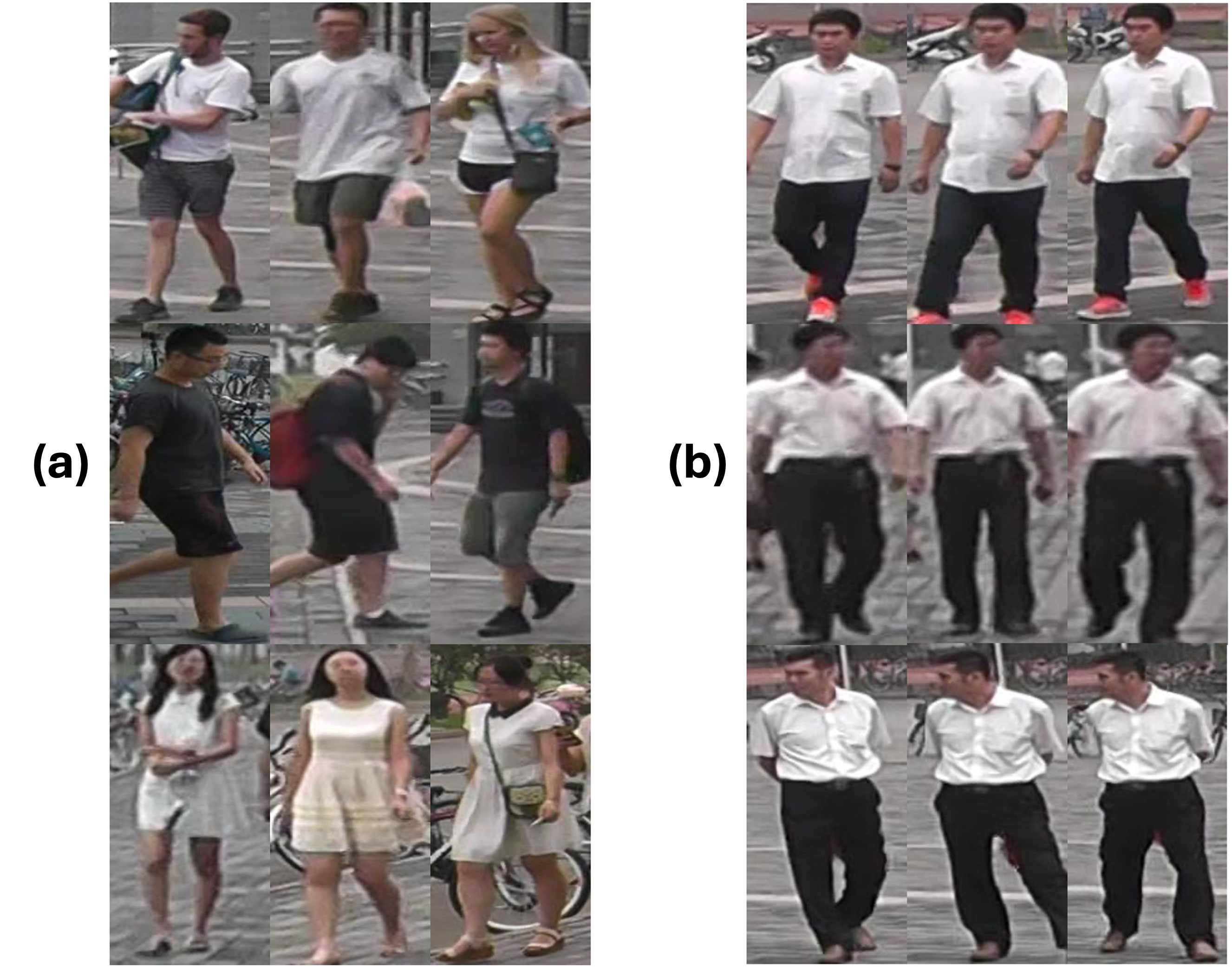}
  \vspace{-5mm}
  \caption{Examples of different identities with similar looks sampled from MARS dataset~\cite{zheng2016mars}. (a) Single frames sampled from different identities that have subtle differences in appearance. Each row consists of three different identities. (b) Multiple frames sampled from three identities that look similar but have subtle differences in spatial and temporal aspects. Each row contains one identity.}
  \label{fig:finegrained}
\vspace{-4mm}
\end{wrapfigure}

Particularly, in real-world surveillance scenarios, different identities often exhibit highly similar global appearance, making coarse-level features insufficiently discriminative. As illustrated in Fig.\ref{fig:finegrained}, such cases can easily confuse existing models. In contrast, fine-grained cues, such as hair, shoes, backpacks, and shoulder bags, can provide critical evidence for distinguishing identities\cite{DenseIL,aich2021factorization}. For example, in Fig.~\ref{fig:finegrained}(a), the persons in white shirts can be distinguished by their \textit{hair} and \textit{bags}, while the persons in black shirts can be differentiated by their \textit{backpacks}.

Many approaches have been proposed to enhance fine-grained feature learning for video-based Re-ID~\cite{DenseIL,aich2021factorization,BiCnet_TKS,liu2021watching,yang2020spatial,zhang2020multi}. For example, several studies incorporate attention mechanisms to highlight informative regions~\cite{DenseIL,BiCnet_TKS,zhang2020multi,liu2021watching}, while others exploit part-level cues, such as attributes and human semantic parsing, to capture fine-grained details~\cite{zhu2022pass,zhu2021aaformer,zhu2020identity}. Though these methods have shown promising results, they still have limitations. In particular, network parameters are typically optimized using all training samples jointly. Yet, as shown in Fig.\ref{fig:finegrained}, there can be various fine-grained cues to discriminate similar samples, each of which may only appear in a small subset of samples.
As a result, optimization over all samples tends to bias the model toward generalizable coarse patterns, rather than learning the subtle cues that might only apply to a small group of samples~\cite{johnson2019survey}.

Therefore, in order to improve the network's ability to learn the fine-grained cues, inspired by specialization learning mechanisms \cite{li2022dynamic,foo2022era}, we develop an \textit{input-aware extendable expert module} that adopts specialized parameters to particularly handle different subsets of similar samples.
More concretely, we construct the expert module with a set of \textit{experts}, each capable of learning information from the input samples.
In this module, we apply an \textit{input-aware expert selection mechanism} to activate the most relevant expert within a set of candidates, which is achieved by evaluating the relevance scores of each expert to the input with convolution operations. 
{Naturally, input samples that are similar tend to have similar convolution results (i.e., expert relevance scores) and thus tend to activate the same experts~\cite{li2022dynamic,foo2022era,zeiler2014visualizing}.
In this way, the experts, activated with similar inputs, only need to focus on learning the fine-grained differences within the particular subsets of similar samples, instead of being pushed to learn more general patterns that apply to more samples.
Hence, each expert gains specialized power to exploit the subtle differences within a subset of similar samples, leading to improved performance in discriminating them.}
Moreover, we design an extendable scheme that allows the expert module to automatically add new experts during training.
Intuitively, having too many experts in the expert module can be inefficient, and expert module of a small size may not be capable of handling large-scale datasets.
With the extendable scheme, the expert module can automatically expand to fit its need, avoiding the trouble of finding a proper number of experts by handcrafted assigning or exhaustive search.

In addition, {in video Re-ID}, fine-grained information may lie in both spatial and temporal aspects. 
We may differentiate some samples using spatial fine-grained cues (e.g., shoes), while temporal information (e.g., gait and temporal pose information) can be useful to distinguish some other samples.
As these subtle cues can exist more in either spatial or temporal aspect for different samples and may be easily neglected, it is crucial for the network to effectively learn and adaptively utilize them to discriminate different samples.

To achieve this goal, motivated by the spatial-temporal specialization in \cite{li2022dynamic}, we adopt the \textit{input-aware spatial-temporal selection mechanism} in each expert to effectively and dynamically exploit the fine-grained differences in spatial and temporal aspects.
After the expert is activated, we explicitly force each parameter to focus on either spatial or temporal aspect.
As the expert is activated on a subset of similar samples with only fine-grained differences, this mechanism will further push each parameter to be more sensitive to subtle differences in a specific aspect.
Meanwhile, the spatial-temporal selection forces the expert to dynamically focus on either spatial or temporal aspect for each input feature channel, allowing the experts to adaptively attach more importance to either spatial or temporal aspects for different samples.
In this way, our expert module {can effectively} exploit fine-grained spatial-temporal information and achieve dynamic spatial-temporal feature learning.

In summary, our contributions are as follows:
(1) To fully exploit fine-grained features for video-based Re-ID, we propose a flexible expert module. We force the {experts} to exclusively discriminate among subsets of similar input samples with an expert selection mechanism, and encourage the experts to learn fine-grained information.
(2) To further effectively enhance spatial-temporal feature learning, we dynamically utilize spatial and temporal fine-grained features based on each input sample with a spatial-temporal selection scheme.
(3) We show the effectiveness of our method by evaluating our method on MARS~\cite{zheng2016mars} and LS-VID~\cite{lsvid}. 
With the help of the input-aware extendable expert module, we are able to reach {outstanding} performances on both datasets.

\section{Related Work}

\textbf{Fine-grained feature learning.}
Fine-grained feature learning has been a longstanding problem. While humans can learn not only the significant information but also the minor details of a sample, it is still a challenge for neural networks. 
The ability to learn fine-grained information is essential to tasks such as fine-grained image classification~\cite{Yang_2022_CVPR,Yang_2022_CVPR1,Touvron_2021_ICCV} and fine-grained action recognition~\cite{li2022dynamic,Munro_2020_CVPR,foo2022era}, where the inter-class differences can be subtle.
In person Re-ID, learning fine-grained features is also an important challenge, especially when differentiating different identities in similar appearance~\cite{he2021transreid,hou2020temporal,hong2021fine,DenseIL,BiCnet_TKS,aich2021factorization} where commonly-used features in coarse level (e.g., color of clothes) may not be discriminative enough. Previous works in exploring fine-grained features can be roughly summarized into two categories: attention-based methods~\cite{DenseIL,BiCnet_TKS,aich2021factorization,zhang2020multi} and part-level methods~\cite{zhang2020multi,yang2020spatial,zhu2022pass,zhu2021aaformer,zhu2020identity}. The attention-based methods train the network to focus on the most informative regions. 
For example, He et al.~\cite{DenseIL} propose Dense Attention that extracts hybrid information from both convolution neural network (CNN) and self-attention to learn the preference of fine-grained features.
Hou et al.~\cite{BiCnet_TKS} propose to learn fine-grained features by applying several parallel attention modules that force the network to explore different regions in the input video sequence. On the other hand, part-level methods explore fine-grained features by leveraging local regions (e.g., sub-parts of the frames and body parts). For example, Zhu et al.~\cite{zhu2022pass} generate part-level features with a knowledge distillation paradigm that learns from both global and part-level patches. Zhu et al. ~\cite{zhu2020identity} propose to learn human body parts and their belongings by human semantic parsing method.

While these methods are all insightful and achieve good performances, they may not be able to {effectively} explore the fine-grained information.
{As the parameters of networks are updated using every samples in the dataset, the loss will push them to learn general patterns to contribute to discriminating more identities, 
as opposed to the fine-grained cues that may only apply to subsets of samples~\cite{johnson2019survey}.
Hence, from a new perspective, we intend to alleviate this problem by explicitly activating the experts on subsets of similar samples and pushing them to only focus on learning the subtle differences within the subsets of similar samples. 
In this way, our expert module gains enhanced ability to effectively learn the fine-grained differences between similar samples.}

\textbf{Spatial-temporal feature learning.}
In video-based Re-ID, leveraging spatial and temporal features is of great significance. 
There are different approaches
~\cite{liu2021watching,BiCnet_TKS,aich2021factorization,yang2020spatial,hou2020temporal,liu2021spatial,wang2021pyramid,eom2021video,yan2020learning,wu2022temporal} to emphasize the importance of utilizing spatial and temporal information. For example, Aich et al.~\cite{aich2021factorization} propose spatial-temporal representation factorization to aggregate temporal and spatial features in high and low frequency. Eom et al.~\cite{eom2021video} introduce spatial and temporal memories to store spatial distractors and typical temporal patterns. Wang et al.~\cite{wang2021pyramid} adopt a pyramid structure that progressively aggregates spatial and temporal features. Also, \cite{yang2020spatial,liu2021spatial} utilize graph convolutions to model the spatial-temporal information between frames.
{Different from previous methods, in this paper, we aim to improve the network's ability to learn fine-grained differences in spatial and temporal aspects by explicitly forcing each of the parameters to focus on one particular aspect.
Because the expert only needs to focus on the fine-grained differences within a subset of similar samples, this mechanism will further push the parameters to be more sensitive to fine-grained cues in spatial and temporal aspects.
Also, the dynamic selection allows the expert to adaptively exploit spatial and temporal information for different samples.}

\textbf{Dynamic network.}
As opposed to static networks, dynamic neural networks can adjust their structures or parameters according to the input, which adds to the representation power, adaptiveness, and interpretability~\cite{han2021dynamic}. Typical dynamic structures include dynamic depth~\cite{wang2018skipnet}, dynamic width~\cite{cai2021dynamic}, and dynamic routing~\cite{hinton2018matrix}. 
Specifically, some previous works \cite{li2022dynamic,foo2022era} generate decisions for dynamic routing using expert modules with fixed structure to learn different human actions. Differently, we design an extendable expert module that can grow its capacity during training, enabling flexible and input-aware modeling of fine-grained spatial and temporal features. Our designs are particularly effective for distinguishing visually similar individuals in video-based person Re-ID.

\section{Method}

\begin{figure*}[t]
  \centering
   \includegraphics[width=0.85\linewidth]{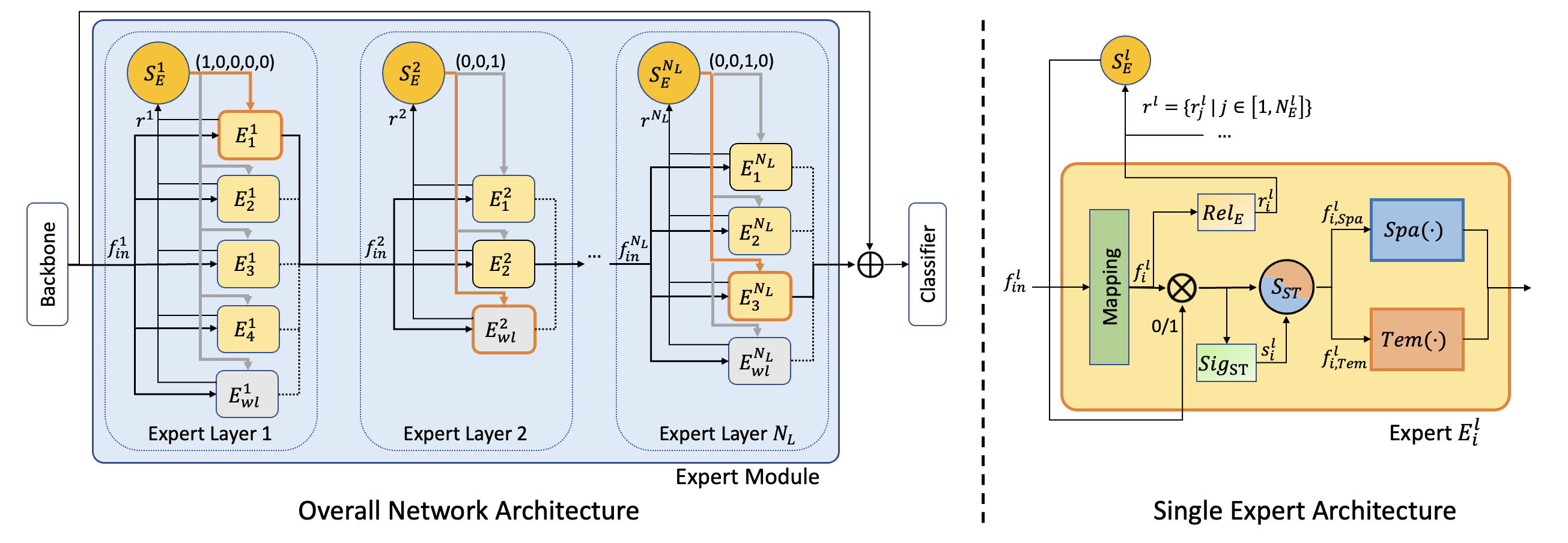}
   \vspace{-3mm}
   \caption{Architecture of the proposed expert module. 
   (Left) Each expert layer contains a number of experts $E_i^l$ and an additional wait-list expert $E_{wl}^l$. For every input feature $f_{in}^l$, the expert selection mechanism $S_E^l$ outputs a one-hot vector according to the relevance score vector $r^l$ and activates one expert in each layer (indicated by orange arrows) while other experts are deactivated (indicated by gray arrows). Then the activated expert produces the output feature of the layer. A skip connection is then applied    (indicated by $\bigoplus$). The classifier is only utilized during training.
   (Right) In the expert $E_i^l$, we first process the input feature $f_{in}^l$ with a mapping procedure and obtain the mapped feature $f_i^l$. Then $f_i^l$ is fed into the relevance evaluation module $\mathrm{Rel}_E$ to compute a relevance score $r_i^l$.
   The expert selector $S_E^l$ in each layer takes relevance scores from every expert and produces a one-hot vector to activate or deactivate the expert (indicated by multiplying 0 or 1 with $\bigotimes$).
   Once the expert is activated, it further computes a spatial-temporal significance vector $s_i^l$ with the spatial-temporal significance module $\mathrm{Sig}_{ST}$. The spatial-temporal selector $S_{ST}$ divides $f_i^l$ into $f_{i,Spa}^l$ and $f_{i,Tem}^l$ according to $s_i^l$, and feed them into spatial and temporal branch respectively. We indicate spatial and temporal branch by $\mathrm{Spa(\cdot)}$ and $\mathrm{Tem(\cdot)}$.} 
   \label{fig:architecture}
   \vspace{-5mm}
\end{figure*}


\subsection{Overview}

One major challenge in video-based person Re-ID is distinguishing different identities with highly similar appearance~\cite{DenseIL,aich2021factorization}. In such cases, coarse appearance cues, such as overall shape and clothing color, are often insufficient for reliable matching. However, most existing methods train network parameters using all samples in the dataset, which tends to encourage the learning of broadly shared patterns rather than subtle fine-grained differences that may only appear in small subsets of visually similar samples.

To address this issue, inspired by specialization learning in fine-grained modeling \cite{li2022dynamic,foo2022era}, we propose an expert module that specifically learns fine-grained discrimination within subsets of \textit{similar samples}. As shown in Fig.~\ref{fig:architecture}, the module contains $N_L$ expert layers, each consisting of $N_E^l$ experts and an expert selector ($l\in[1,N_L]$). Given the coarse-grained features extracted by a backbone network, the selector dynamically activates the most relevant expert in each layer according to the input. Since similar samples tend to yield similar relevance scores, they are likely to activate the same experts. As a result, each expert is updated on a subset of similar samples, encouraging it to capture subtle discriminative cues. To avoid manually specifying the number of experts, we further introduce an extendable scheme that automatically appends wait-list experts when needed.
In addition, to better exploit fine-grained video cues, inspired by \cite{li2022dynamic}, we adopt parallel spatial and temporal branches in each activated expert. An input-aware spatial-temporal feature selection mechanism dynamically routes each feature channel to either the spatial or temporal branch, allowing the model to adaptively focus on the most informative subtle differences for each sample. 

With these two dynamic selection mechanisms, the proposed module can effectively learn fine-grained spatial-temporal representations for video-based person Re-ID. Below we formally introduce the details of the proposed mechanisms.

\subsection{Input-aware Extendable Expert Selection Mechanism}\label{sec:expert_selection}

In each expert layer, we adopt a selection mechanism to dynamically match the input with the most relevant expert \cite{li2022dynamic,foo2022era}. 
Also, to flexibly construct the expert module to handle the samples, we design an extendable scheme that allows adding new experts during training. 

More concretely, given a video clip, we first extract the coarse-grained features with the backbone network as the input to the expert module.
We denote the input feature to the $l$-th expert layer as $f_{in}^l \in \mathbb{R}^{C\times T \times H \times W}$.
In the $i$-th expert $E_i^l$, we process the input feature with the mapping module and obtain the mapped feature $f_i^l \in \mathbb{R}^{C\times T \times H \times W}$.
Then the mapped feature $f_i^l$ is fed into the relevance evaluation module $\mathrm{Rel}_E$ to evaluate the relevance of the expert to the input sample.
Specifically, in $\mathrm{Rel}_E$, we adopt max-pooling and a fully connected layer to obtain a relevance value, and normalize the value with $\tanh$ function to obtain the relevance score $r_i^l\in \mathbb{R}$:
\begin{equation}
\setlength{\abovedisplayskip}{2pt}
\setlength{\belowdisplayskip}{2pt}
 \label{eq:score}
\begin{aligned}
    f_i^l &= \mathrm{Mapping}(f_{in}^l),\\
    {f}_{i,\max}^l & = \max_{T, H, W}(f_i^l),\ {f}_{i,\max}^l\in \mathbb{R}^{C\times 1}, \\
    r_i^l & = \tanh(w_E^{i,l} {f}_{i,\max}^l), w_E^{i,l}\in \mathbb{R}^{1\times C},
\end{aligned}
\end{equation}
where $\mathrm{Mapping}(\cdot)$ represents the convolution operation in the mapping module, and $w_E^{i,l}$ denotes the linear transformation in the fully connected layer.

After obtaining relevance scores for every expert in the layer, the aggregated relevance vector $r^l = \{r_1^l, r_2^l, ..., r_{N_E^l}^l\}$ is fed into the expert selector $S_E^l$.
Using the Gumbel-Softmax method~\cite{jang2017categorical,li2022dynamic,foo2022era}, $S_E^l$ evaluates $r^l$ and generates a one-hot vector with the $s$-th value being 1.
We then activate the $s$-th expert $E_s^l$ and deactivate other experts in the layer. 
Notably, only the activated expert $E_s^l$ is involved in the following process,
and therefore during training, the parameters of each expert are only updated when activated.

In addition, we adopt an extendable expert scheme by keeping an additional expert in each layer (termed as wait-list expert $E_{wl}^l$) during training. For each input sample, the wait-list expert also generates a relevance score, and
is only added to the layer if it has a higher relevance score to the input samples than the existing ones. 
Once the expert $E_{wl}^l$ is appended to layer $l$, it becomes a constant expert $E_{N_L+1}^l$ and a new wait-list expert is generated automatically.
The intuition behind this extendable scheme is that it empowers the network to adaptively adjust the number of experts and avoids the trouble of manually assigning the proper number of experts.

The expert selection mechanism plays an important role in improving the ability to learn fine-grained information. 
{Most significantly, it achieves dynamic activation of the experts on subsets of similar samples.
As the experts are only updated on subsets of similar samples instead of being trained with all samples, they are forced to focus on learning fine-grained differences within particular subsets, which promotes the ability to discriminate similar samples.}

\subsection{Input-aware Spatial-Temporal Selection Mechanism}\label{sec:st_selection}
After activating the expert for the input sample, we further explore spatial and temporal fine-grained features dynamically. 
Specifically, inspired by spatial-temporal fine-grained feature modeling \cite{li2022dynamic}, we construct two parallel branches that focus on spatial and temporal information respectively as shown in Fig.\ref{fig:architecture}. 
The spatial branch consists of a $1\times3\times3$ convolution layer to express spatial information, while the temporal branch contains a $3\times1\times1$ convolution layer that learns temporal information. 
{Motivated by the observation that the discriminative information may lie more in either spatial or temporal aspects for different samples, we allow the experts to adaptively adjust their emphasis on the two aspects by forcing every feature channel to dynamically select between spatial and temporal branches for each input sample.}

More concretely, we leverage the mapped feature $f_s^l$ in the selected expert $E_s^l$ and compute a significance vector $s_s^l\in \mathbb{R}^{C\times 1}$ with the spatial-temporal significance evaluation module $\mathrm{Sig}_{ST}$,
indicating whether the feature channel is more significant in spatial aspect or temporal aspect.
In $\mathrm{Sig}_{ST}$, we utilize max-pooling and fully connected layer to generate a vector $s_s^l\in \mathbb{R}^{C}$ that contains one significance value for each feature channel:
\begin{equation}
\setlength{\abovedisplayskip}{2pt}
\setlength{\belowdisplayskip}{2pt}
 \label{eq:st_significance}
\begin{aligned}
    {f}_{s,\max}^l & = \max_{T, H, W}(f_s^l), {f}_{s,\max}^l\in \mathbb{R}^{C\times 1}, \\
    s_s^l & = \tanh(w_{ST}^{s,l} {f}_{s,\max}^l), w_{ST}^{s,l}\in \mathbb{R}^{C\times C},
\end{aligned}
\end{equation}
where $w_{ST}^{s,l}$ represents the linear transformation in the fully connected layer.

After obtaining the significance vector $s_s^l$, we generate a binary decision vector $d_{s}^{l}\in \mathbb{R}^{C}$ with the help of the Improved Semhash method~\cite{kaiser2018fast,kaiser2018discrete}. 
Specifically, for the $c$-th feature channel, we obtain a decision value $d_{s,c}^{l}$. When $d_{s,c}^{l}=1$, the feature slice $f_{s,c}^l\in \mathbb{R}^{1\times T\times H\times W}$ goes to the spatial branch; otherwise, it is fed into the temporal branch.
The selection process is formulated as:
\begin{equation}
\setlength{\abovedisplayskip}{2pt}
\setlength{\belowdisplayskip}{2pt}
\begin{aligned}
    f_{s,Spa}^l &= f_{s}^l \odot d_{s}^{l}, \quad
    f_{s,Tem}^l &= f_{s}^l \odot (\textbf{1}-d_{s}^{l}), \\
\end{aligned}
\label{eq:decision}
\end{equation}
where $f_{s,Spa}^l$ and $f_{s,Tem}^l$ represent the input features to spatial and temporal branches respectively, $\textbf{1}$ is a vector of 1's of size $C$, $\odot$ denotes multiplication in the channel dimension, and the elements in $d_{s}^{l}$ and $(\textbf{1}-d_{s}^{l})$ are considered as channels for simplicity. 

The output feature $f_{s,out}^l$ of the $s$-th expert in the $l$-th layer is obtained by adding the outputs from the two branches:
\begin{equation}
\setlength{\abovedisplayskip}{2pt}
\setlength{\belowdisplayskip}{2pt}
    f_{s,out}^l = \mathrm{Spa}(f_{s,Spa}^l) + \mathrm{Tem}(f_{s,Tem}^l),
\end{equation}
where $\mathrm{Spa(\cdot), Tem(\cdot)}$ represent operations in the spatial and temporal branches respectively.

{As the expert is forced to choose between spatial and temporal branches for each input feature channel, during training, it learns to adaptively assign the feature channels to the branch that can lead to greater discriminative capacity.
In this way, the experts are able to effectively and efficiently explore fine-grained information in spatial and temporal aspects according to the input samples.}
By forcing the spatial and temporal branches to focus on their specialties, they are stimulated to be sensitive to subtle differences in {their own specialties}.
Therefore, during inference, the experts are able to dynamically utilize spatial and temporal features and effectively exploit fine-grained information. 

\subsection{Loss Function}
In our expert module, we encourage the experts to learn fine-grained features to identify different identities. Intuitively, the experts in the same layer need to be less similar to each other in order to handle different samples in the dataset. 
To achieve this goal, we additionally adopt a diversity loss $\mathcal{L}_{div}$ to limit the pair-wise similarity among the experts.
More concretely, for each expert, we first obtain $p_i^l$ and $q_i^l$ by vectorizing the parameters in the spatial and temporal branches respectively, and compute the pair-wise cosine similarity of the vectorized spatial and temporal parameters within the same layer:
\begin{equation}
\begin{aligned}
\setlength{\abovedisplayskip}{2pt}
\setlength{\belowdisplayskip}{2pt}
    \mathcal{L}_{div}^{l,Spa} &= \sum_{i=1}^{N_{E}}\sum_{j=1, j \neq i}^{N_{E}}{\frac{{{p}_i^l}^\top  {p}_j^l}{\lVert {p}_i^l \rVert_2 \lVert {p}_j^l \rVert_2}}, \quad
    \mathcal{L}_{div}^{l,Tep} &= \sum_{i=1}^{N_{E}}\sum_{j=1, j \neq i}^{N_{E}}{\frac{{{q}_i^l}^\top  {q}_j^l}{\lVert {q}_i^l \rVert_2 \lVert {q}_j^l \rVert_2}}. 
\end{aligned}
\label{eq:l_div_1}
\end{equation}

We further aggregate the diversity loss across all layers by summation:
\begin{equation}
\setlength{\abovedisplayskip}{2pt}
\setlength{\belowdisplayskip}{2pt}
    \label{eq:l_div_2}
    \mathcal{L}_{div} = \frac{1}{2N_L}\sum_{l=1}^{N_L}\mathcal{L}_{div}^{l,Spa} + \frac{1}{2N_L}\sum_{l=1}^{N_L}\mathcal{L}_{div}^{l,Temp}.
\end{equation}

In this way, a small $\mathcal{L}_{div}$ indicates that the parameters of the experts are nearly orthogonal to each other, so they share minimal common knowledge. During training, the diversity loss will penalize the pair-wise similarity between the experts, which helps the network to exploit diverse fine-grained information.

Overall, we train the network with the combination of cross entropy loss $\mathcal{L}_{ce}$, batch hard triplet loss $\mathcal{L}_{tri}$~\cite{triplet_loss}, and $\mathcal{L}_{div}$:
\begin{equation}
\setlength{\abovedisplayskip}{2pt}
\setlength{\belowdisplayskip}{2pt}
    \mathcal{L} = \mathcal{L}_{ce} + \mathcal{L}_{tri} + \lambda \mathcal{L}_{div},
    \label{eq:loss}
\end{equation}
where $\lambda$ is a hyperparameter to balance the influence of diversity loss.

\section{Experiments}
\textbf{Datasets.} We carry out experiments to evaluate the performance of our proposed methods on two large-scale video-based person Re-ID datasets: MARS~\cite{zheng2016mars} and LS-VID~\cite{li2019global}.
Mars~\cite{zheng2016mars} is one of the largest video-based person Re-ID datasets, consisting of 17,503 sequences captured by six cameras. There are 1,261 identities in total, with 625 identities in the training set and 636 identities in the test set. 
LS-VID~\cite{li2019global} dataset is another large-scale dataset for video-based person Re-ID captured by 3 indoor cameras and 12 outdoor cameras. This dataset consists of 14,943 sequences of 3,772 pedestrians. There are 842 identities in the training set, 200 identities in the validation set, and 2,730 identities in the test set.

\textbf{Evaluation Metrics.}
Following previous person Re-ID methods~\cite{bai2022salient,DenseIL,BiCnet_TKS,wang2021pyramid}, we adopt Cumulated Matching Characteristics (CMC) curve and mean Average Precision (mAP) as evaluation metrics. 

\subsection{Implementation Details}
Following SINet~\cite{bai2022salient}, we use ResNet-50~\cite{resnet50} as the backbone.
The expert module is initialized with two experts ($N_E^l=2$) for three layers ($N_L=3$). 
In each expert, the mapping module consists of a $1\times 1\times 1$ convolution layer, followed by batch normalization and ReLU.
The spatial branch contains a $1\times 3\times 3$ convolution, batch normalization and ReLU, while the temporal branch consists of a $3\times 1\times 1$ convolution, followed by batch normalization and ReLU.
We apply $1\times 1\times 1$ convolution to reduce the feature channels before appending the expert module.
The dimensions of the features ($C, T, H, W$) are dependent on the backbone network.
We set the hyperparameter $\lambda$ in Eq.\ref{eq:loss} to 0.1.
During training, we adopt the Restricted Random Sampling (RRS) strategy~\cite{liu2019spatially} and sample 4 frames for each tracklet. Following~\cite{bai2022salient,wang2021pyramid}, the training batch is constructed with 8 different identities, each including 4 tracklets.
The network is trained on a single RTX 3090 GPU. We use the Adam optimizer with learning rate of 0.0005 and weight decay of 0.0005. 
During inference, we split the videos into clips of 4 frames and adopt cosine similarity to measure the distance between the query and the gallery. 
We add noises sampled from standard Gaussian distribution to Improved Semhash during training, and no noises are added during inference. The temperature in Gumbel-Softmax~\cite{jang2017categorical} is set to 1.

\begin{table}[t]
\begin{center}
\caption{Comparison of our method with state-of-the-art video-based person Re-ID methods on MARS and LS-VID.}
\doublerulesep=0.5pt
\resizebox{0.45\textwidth}{!}
{
\begin{tabular}{|c|c|c|c|c|c|c|c|c|c|c|c|c|c|c|c|c|c|c|c|c|c|c|c|c|c|c|c|c|c|c|c|c|c|c|c|c|c|c|c|c|c|c|c|c|c|c|c|c|c|c|c|c|c|c|c|c|c|c|c|c|c|c|c|c|c|c|c|c|c|c|c|c|c|c|c|c|c|c|c|c|c|c|c} 
\hline
\multicolumn{8}{c|}{}
&\multicolumn{16}{c|}{MARS}
&\multicolumn{8}{c}{LS-VID}
\\
\multicolumn{8}{c}{Methods}
&\multicolumn{4}{|c}{mAP}
&\multicolumn{4}{c}{rank-1}
&\multicolumn{4}{c}{rank-5}
&\multicolumn{4}{c|}{rank-20}
&\multicolumn{4}{c}{mAP}
&\multicolumn{4}{c}{rank-1}
\\
\hline
\multicolumn{8}{l|}{MGRA~\cite{zhang2020multi}}
&\multicolumn{4}{c}{85.9}
&\multicolumn{4}{c}{88.8}
&\multicolumn{4}{c}{97.0}
&\multicolumn{4}{c}{{98.5}}
&\multicolumn{4}{|c}{-}
&\multicolumn{4}{c}{-}
\\
\multicolumn{8}{l|}{STGCN~\cite{yang2020spatial}}
&\multicolumn{4}{c}{83.7}
&\multicolumn{4}{c}{89.9}
&\multicolumn{4}{c}{-}
&\multicolumn{4}{c}{-}
&\multicolumn{4}{|c}{-}
&\multicolumn{4}{c}{-}
\\
\multicolumn{8}{l|}{TCLNet-tri~\cite{hou2020temporal}}
&\multicolumn{4}{c}{85.1}
&\multicolumn{4}{c}{89.8}
&\multicolumn{4}{c}{-}
&\multicolumn{4}{c}{-}
&\multicolumn{4}{|c}{-}
&\multicolumn{4}{c}{-}
\\
\multicolumn{8}{l|}{BiCnet-TKS~\cite{BiCnet_TKS}}
&\multicolumn{4}{c}{86.0}
&\multicolumn{4}{c}{90.2}
&\multicolumn{4}{c}{-}
&\multicolumn{4}{c}{-}
&\multicolumn{4}{|c}{75.1}
&\multicolumn{4}{c}{84.6}
\\
\multicolumn{8}{l|}{GRL~\cite{liu2021watching}}
&\multicolumn{4}{c}{84.8}
&\multicolumn{4}{c}{{91.0}}
&\multicolumn{4}{c}{96.7}
&\multicolumn{4}{c}{98.4}
&\multicolumn{4}{|c}{-}
&\multicolumn{4}{c}{-}
\\
\multicolumn{8}{l|}{CTL~\cite{liu2021spatial}}
&\multicolumn{4}{c}{{86.7}}
&\multicolumn{4}{c}{{91.4}}
&\multicolumn{4}{c}{96.8}
&\multicolumn{4}{c}{{98.5}}
&\multicolumn{4}{|c}{-}
&\multicolumn{4}{c}{-}
\\
\multicolumn{8}{l|}{STRF~\cite{aich2021factorization}}
&\multicolumn{4}{c}{86.1}
&\multicolumn{4}{c}{90.3}
&\multicolumn{4}{c}{-}
&\multicolumn{4}{c}{-}
&\multicolumn{4}{|c}{-}
&\multicolumn{4}{c}{-}
\\
\multicolumn{8}{l|}{DenseIL~\cite{DenseIL}}
&\multicolumn{4}{c}{{87.0}}
&\multicolumn{4}{c}{90.8}
&\multicolumn{4}{c}{{97.1}}
&\multicolumn{4}{c}{{98.8}}
&\multicolumn{4}{|c}{-}
&\multicolumn{4}{c}{-}
\\
\multicolumn{8}{l|}{STMN~\cite{eom2021video}}
&\multicolumn{4}{c}{84.5}
&\multicolumn{4}{c}{90.5}
&\multicolumn{4}{c}{-}
&\multicolumn{4}{c}{-}
&\multicolumn{4}{|c}{69.2}
&\multicolumn{4}{c}{82.1}
\\
\multicolumn{8}{l|}{PSTA~\cite{wang2021pyramid}}
&\multicolumn{4}{c}{85.8}
&\multicolumn{4}{c}{{91.5}}
&\multicolumn{4}{c}{-}
&\multicolumn{4}{c}{-}
&\multicolumn{4}{|c}{-}
&\multicolumn{4}{c}{-}
\\
\multicolumn{8}{l|}{SINet~\cite{bai2022salient}}
&\multicolumn{4}{c}{86.2}
&\multicolumn{4}{c}{{91.0}}
&\multicolumn{4}{c}{-}
&\multicolumn{4}{c}{-}
&\multicolumn{4}{|c}{{79.6}}
&\multicolumn{4}{c}{{87.4}}
\\
\multicolumn{8}{l|}{CAViT~\cite{wu2022cavit}}
&\multicolumn{4}{c}{\textbf{87.2}}
&\multicolumn{4}{c}{{90.8}}
&\multicolumn{4}{c}{-}
&\multicolumn{4}{c}{-}
&\multicolumn{4}{|c}{{79.2}}
&\multicolumn{4}{c}{\textbf{89.2}}
\\
\multicolumn{8}{l|}{DSANet~\cite{kim2023feature}}
&\multicolumn{4}{c}{86.6}
&\multicolumn{4}{c}{{91.1}}
&\multicolumn{4}{c}{-}
&\multicolumn{4}{c}{-}
&\multicolumn{4}{|c}{{75.5}}
&\multicolumn{4}{c}{{85.1}} 
\\
\multicolumn{8}{l|}{MIRE-GRAR~\cite{zhu2025multi}}
&\multicolumn{4}{c}{86.6}
&\multicolumn{4}{c}{{91.5}}
&\multicolumn{4}{c}{96.8}
&\multicolumn{4}{c}{98.5}
&\multicolumn{4}{|c}{{75.5}}
&\multicolumn{4}{c}{{85.1}} 
\\
\multicolumn{8}{l|}{EGMDL~\cite{cao2026learning}}
&\multicolumn{4}{c}{86.6}
&\multicolumn{4}{c}{{91.1}}
&\multicolumn{4}{c}{-}
&\multicolumn{4}{c}{-}
&\multicolumn{4}{|c}{{-}}
&\multicolumn{4}{c}{{-}} 
\\
\hline
\multicolumn{8}{l|}{Ours}
&\multicolumn{4}{c}{{87.0}}
&\multicolumn{4}{c}{\textbf{91.6}}
&\multicolumn{4}{c}{\textbf{97.4}}
&\multicolumn{4}{c}{\textbf{98.9}}
&\multicolumn{4}{|c}{\textbf{81.0}}
&\multicolumn{4}{c}{{88.3}}
\\
\hline
\end{tabular}
}
\label{tab:sota}
\vspace{-5mm}
\end{center}
\end{table}

\subsection{Performance Comparison}
We compare the performance of our proposed method with state-of-the-art methods~\cite{gu2019temporal,zhang2020multi,yang2020spatial,hou2020temporal,BiCnet_TKS,liu2021watching,liu2021spatial,aich2021factorization,DenseIL,eom2021video,wang2021pyramid,bai2022salient} on two large-scale datasets: MARS~\cite{zheng2016mars} and LS-VID~\cite{lsvid}. The results are shown in Tab.\ref{tab:sota}. 
On both datasets, we initialize the three expert layers with two experts and one wait-list expert, and allow the expert module to automatically append new experts to each layer. 
For MARS, the expert module stabilizes with expert number of $N_E^1, N_E^2, N_E^3=4, 2, 4$, while on LS-VID the expert module expands to a size of $N_E^1, N_E^2, N_E^3=4, 4, 4$.

On MARS, our proposed method achieves 87.0\% in mAP and 91.6\% in rank-1, achieving state-of-the-art performance. 
Specifically, compared with other methods exploring fine-grained features~\cite{zhang2020multi,yang2020spatial,BiCnet_TKS,DenseIL,liu2021watching,aich2021factorization}, our method reaches top performance.
This shows that our expert module has superior ability to learn useful information that can help improve performance.
On LS-VID, our expert module {also shows outstanding performance} on both mAP and rank-1 metrics. 
{Although CAViT~\cite{wu2022cavit} also shows competitive performances, we argue that CAViT is based on transformer and samples 8 frames for each identity while our results are achieved based on CNN with 4 frames sampled for each identity due to limitation of GPU memory. }

\subsection{Ablation Studies}

We conduct ablation studies 
to evaluate our designs on LS-VID. 

\setlength{\columnsep}{0.15in}
\begin{wraptable}[5]{r}{0.42\columnwidth}
    \centering
\vspace{-9mm}
\caption{Impact of the expert selection mechanism $S_E$.}
\vspace{-3mm}
\resizebox{\linewidth}{!}
{
\begin{tabular}{@{}l|ccc|cccc@{}}
\hline
Models           & mAP & rank-1 \\ 
\hline
ResNet-50 & 73.3 & 82.8 \\
Expert Module w/o $S_E$ (average weighting) & 79.9 & 87.3 \\ 
Expert Module w/o $S_E$ (random activation) & 79.6 & 87.2 \\ 
Expert Module w/ $S_E$ & {81.0} & {88.3} \\
\hline
\end{tabular}
}
\label{tab:ablation_s_e}
\vspace{-10mm}
\end{wraptable}
\textbf{Expert Selection Mechanism.}
To evaluate the importance of the expert selection mechanism $S_E$, 
we conduct two experiments without the expert selection mechanism as shown in Tab.\ref{tab:ablation_s_e}: 1) activating all experts and averaging the outputs (\textbf{average weighting}), and 2) randomly activating the experts (\textbf{random activation}).
As shown, the performances drop when we disable the expert selection mechanism.
Our model achieves better results than the experiment with \textbf{average weighting} suggesting that our method benefits from {dynamically activating and pushing the experts to specifically focus on subsets of samples compared to simply adding more model parameters through the experts}.
{Also, compared to \textbf{random activation} of the experts, we show that the expert module is more effective with the input-aware design that intends to only activate the experts on subsets of similar samples.}

\setlength{\columnsep}{0.15in}
\begin{wraptable}{r}{0.45\columnwidth}
    \centering
\vspace{-9mm}
\caption{Impact of the spatial-temporal selection mechanism $S_{ST}$.}
\vspace{-3mm}
\resizebox{0.45\textwidth}{!}
{
\begin{tabular}{@{}l|ccc|cccc@{}}
\hline
Models           & mAP & rank-1 \\ 
\hline
Expert Module w/o $S_{ST}$ (spatial branch) & 79.5 & 87.3 \\ 
Expert Module w/o $S_{ST}$ (temporal branch) & 79.4 & 87.1 \\ 
Expert Module w/o $S_{ST}$ (single branch) & 80.0 & 87.4 \\ 
Expert Module w/o $S_{ST}$ (average fusion) & 80.1 & 87.6 \\ 
Expert Module w/o $S_{ST}$ (random selection) & 80.0 & 87.3 \\ 
Expert Module w/ $S_{ST}$ & {81.0} & {88.3} \\
\hline
\end{tabular}
}
\label{tab:ablation_s_st}
\vspace{-8mm}
\end{wraptable}
\textbf{Spatial-temporal selection mechanism.}
We investigate the importance of the spatial-temporal selection mechanism $S_{ST}$ as shown in Tab.\ref{tab:ablation_s_st}.
We evaluate {the following} variants: 
1) {only using the spatial branch or the temporal branch in each expert (\textbf{spatial branch} and \textbf{temporal branch})},
2) replacing the two-branch design with a single branch containing a $3\times 3\times 3$ convolution layer with batch normalization and ReLU (\textbf{single branch}),
3) {fusing the two branches by averaging their outputs (\textbf{average fusion})}, and 
4) {replace the selection mechanism with random assignment} (\textbf{random selection}).
From the results in Tab.\ref{tab:ablation_s_st} we can observe that only learning fine-grained information in one aspect is not sufficient. Also, our expert model outperforms all compared variants, demonstrating its effectiveness.

\setlength{\columnsep}{0.15in}
\begin{wraptable}{r}{0.45\columnwidth}
    \centering
\vspace{-10mm}
\caption{Impact of the diversity loss.}
\vspace{-3mm}
\resizebox{0.3\textwidth}{!}
{
\begin{tabular}{@{}l|ccc|cccc@{}}
\hline
Models           & mAP & rank-1 \\ 
\hline
$\lambda=0$ (without $\mathcal{L}_{div}$) & 80.1 & 87.3 \\ 
$\lambda=0.05$ & 80.6 & 87.9 \\ 
$\lambda=0.1$ & {81.0} & {88.3} \\
$\lambda=0.15$ & 80.3 & 87.7 \\
$\lambda=0.5$ & 80.0 & 87.6 \\ 
\hline
\end{tabular}
}
\label{tab:ablation_loss}
\vspace{-10mm}
\end{wraptable}
\textbf{Diversity loss.}
In our method, we employ the diversity loss $\mathcal{L}_{div}$ to encourage the experts in the same layer to be more diverse. 
To evaluate its impact, we compare the performance of the expert module with diversity loss using different $\lambda$.
As shown in Tab.\ref{tab:ablation_loss}, adding $\mathcal{L}_{div}$ in the training scheme improves rank-1 and mAP scores, and our method achieves the best performance when $\lambda=0.1$. Thus we set $\lambda=0.1$ in our experiments.

\setlength{\columnsep}{0.15in}
\begin{wraptable}{r}{0.2\columnwidth}
    \centering
\vspace{-9mm}
\caption{Number of experts.} 
\vspace{-2mm}
\centering
\resizebox{0.8\linewidth}{!}{
\begin{tabular}{l |c  c }
\hline 
$N_E$  & mAP & rank-1 \\ 
\hline
2 & 80.3 & 87.5\\
3 & 80.4 & 87.7\\
4 & 81.0 & 88.3\\
5 & 80.4 & 87.9\\
\hline
\end{tabular}}
\label{tab:n_expert}
\vspace{-10mm}
\end{wraptable}
\textbf{Extendable scheme.}
To evaluate the design of the extendable scheme which automatically decides the number of experts in each layer, we manually fix the number of experts and compare the performances. 
In these experiments, we disable all wait-list experts and assign the expert module with $N_E=2,3,4,5$ experts per layer. We also keep other settings the same as the main experiment, i.e., we construct 3 layers of experts and enable the selection mechanisms and diversity loss. The results are shown in Tab.\ref{tab:n_expert}.
As we can see, the scores increase when $N_E$ grows from 2 to 4, and the improvement tapers off when $N_E$ exceeds 4.
Notably, the model with expert number automatically assigned with the extendable scheme (i.e., $N_E=4$) surpasses all other models, which shows the effectiveness of the extendable scheme.

\textbf{Number of layers.}
To evaluate the impact of the number of layers in the expert module, we further conduct experiments with different layers of experts. 
\setlength{\columnsep}{0.15in}
\begin{wraptable}{r}{0.25\columnwidth}
    \centering
\vspace{-1mm}
\caption{Number of expert layers.}
    \vspace{-2mm}
    \resizebox{0.63\linewidth}{!}{
    \begin{tabular}{l |c  c }
    \hline 
    $N_L$ & mAP & rank-1 \\ 
    \hline
    1 & 79.7 & 87.3  \\ 
    2 & 80.4 & 87.5 \\
    3 & 81.0 & 87.9 \\
    5 & 80.9 & 87.8 \\
    \hline
    \end{tabular}}
    \label{tab:n_layer}
\vspace{-10mm}
\end{wraptable}
We construct the module with {four} experts per layer, and form the expert module with $N_L=1,2,3,5$ layer(s), as shown in Tab.\ref{tab:n_layer}.
As $N_L$ grows from 1 to 3, the mAP increases by {1.3\%}.
Yet, stacking too many layers (e.g., $N_L=5$) doesn't bring improvement to the performance. 
Thus we set $N_L=3$ in our experiments.

\begin{figure}[t]
  \centering
  \includegraphics[width=0.8\linewidth]{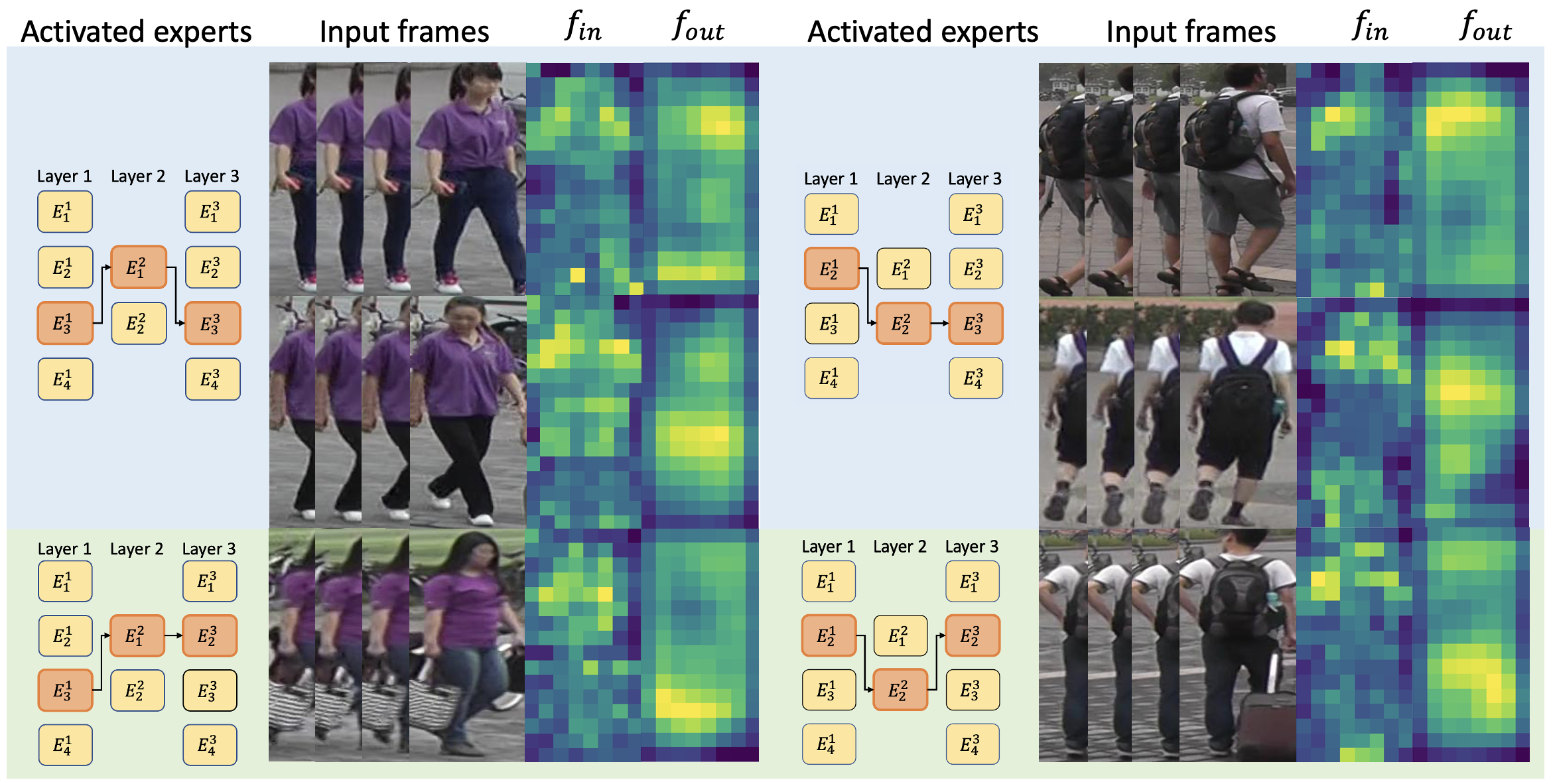}
  \caption{Visualization of samples and features with the activated experts on MARS. The activated experts are indicated in orange, $E_i^l$ represents the $i$-th expert in the $l$-th layer, $f_{in}$ indicates the input feature to the expert module, $f_{out}$ denotes the output feature of the expert module.}
  \label{fig:vis}
  \vspace{-4mm}
\end{figure}

\textbf{Visualizations.}
To better evaluate the design of our expert module, we visualize some examples of the input samples and the activated experts in Fig.\ref{fig:vis}. 
{We show two groups of examples from MARS~\cite{zheng2016mars} with the input frames, input features going into the expert module, and the output features of the expert module.} 
From the visualization results, we can observe that samples that look similar tend to have similar features and activate similar expert sets.
{While the input samples and input features are similar, our expert module can effectively learn discriminative information as shown by the different output features in Fig.\ref{fig:vis}, demonstrating the effectiveness of our design.}

\section{Conclusion}
In this paper, we propose an input-aware extendable expert module to tackle video-based Re-ID problem by exploiting the fine-grained discriminative details. We achieve this by dynamically activating the most relevant experts for the input samples, and adaptively adjusting the spatial-temporal feature learning according to the input sample. In this way, we encourage the network to exploit fine-grained spatial and temporal information. We demonstrate the effectiveness of our method with experiments on two large-scale video-based Re-ID datasets.

\textbf{Privacy and ethical considerations.} Video-based person re-identification relies on surveillance footage, raising privacy concerns. We use only publicly available benchmarks for research purposes and do not identify real-world individuals. Notably, our expert module operates on abstract feature representations rather than raw images, and its modular design is naturally compatible with privacy-preserving techniques such as federated learning across cameras or adversarial feature perturbation to prevent identity leakage. We encourage responsible deployment with appropriate legal safeguards and oversight.

{
    \small
    \bibliographystyle{IEEEtran}
    \bibliography{main}
}
\end{document}